\documentclass[10pt,twocolumn,letterpaper]{article}

\usepackage{iccv}              % To produce the CAMERA-READY version
\definecolor{iccvblue}{rgb}{0.21,0.49,0.74}
\usepackage[pagebackref,breaklinks,colorlinks,allcolors=iccvblue]{hyperref}

\usepackage[dvipsnames,table]{xcolor}
\usepackage{multirow}
\usepackage{pifont}
\usepackage{caption}
\usepackage{cuted}
\usepackage[accsupp]{axessibility}

\newcommand{\ourmethod}{\textsl{attend-and-segment}\xspace}
\newcommand{\ourmodel}{\textsc{DiffLMM}\xspace}

\newcommand{\cmark}{\ding{51}} % check mark
\newcommand{\xmark}{\ding{55}} % cross mark
\definecolor{cellgray}{gray}{0.9}
\definecolor{modelblue}{RGB}{29, 96, 130}
\definecolor{modelorange}{RGB}{233, 113, 49}

\def\paperID{35} % *** Enter the Paper ID here
\def\confName{ICCV Findings}
\def\confYear{2025}

\title{Emergent Visual Grounding in Large Multimodal Models \\ \emph{Without} Grounding Supervision}
\author{%
Shengcao Cao \quad Liang-Yan Gui \quad Yu-Xiong Wang \\
University of Illinois Urbana-Champaign \\
\texttt{\{cao44, lgui, yxw\}@illinois.edu}
}

\begin{document}

\maketitle

\begin{strip}
    \centering
    \vspace*{-12mm}
    \includegraphics[width=\textwidth]{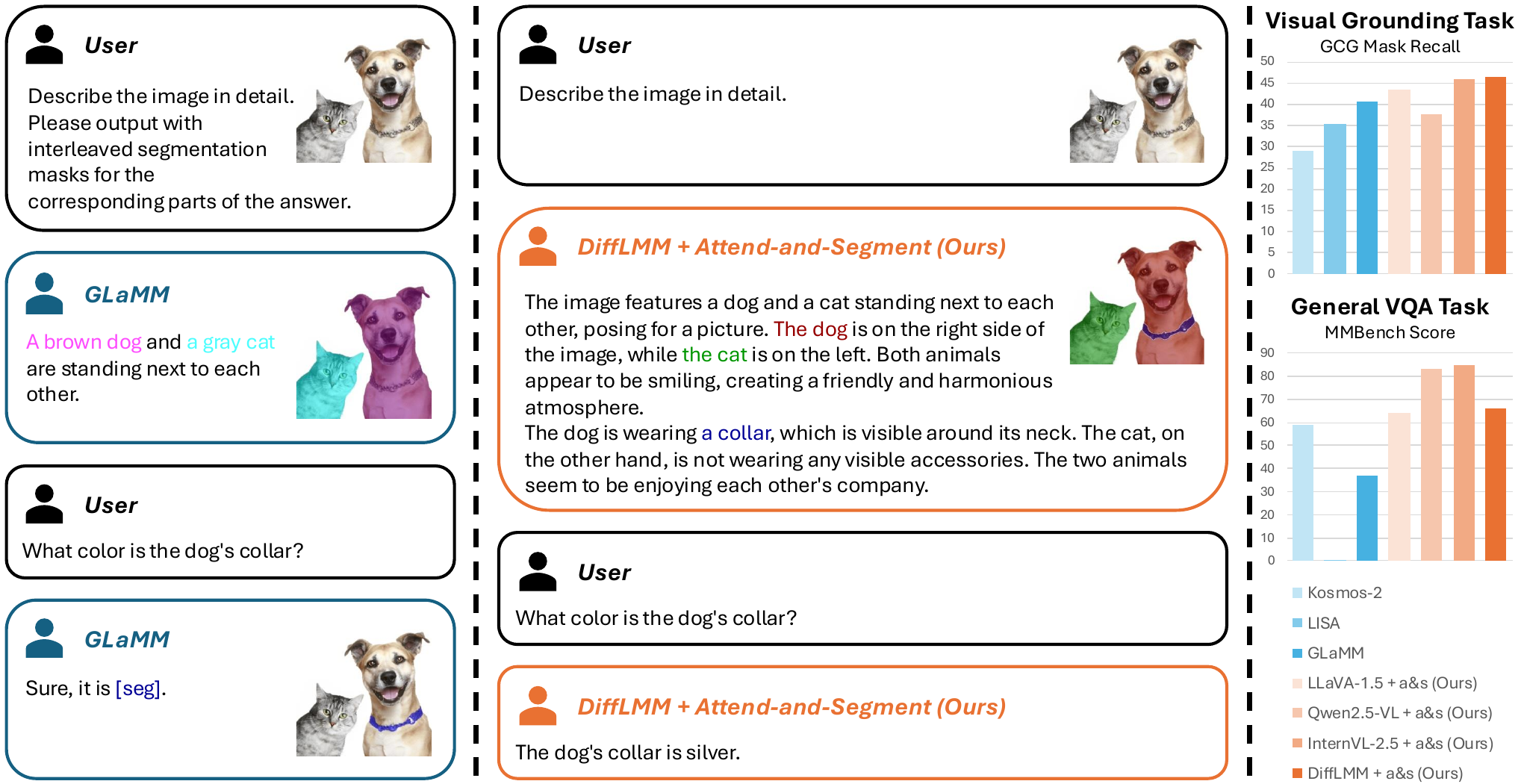}
    \captionof{figure}{\textbf{Grounded conversations with GLaMM~\citep{rasheed2024glamm} \vs our approach, \ourmodel + \ourmethod.} \textbf{Left:} As a state-of-the-art grounding LMM, GLaMM is \emph{trained} to relate text phrases with segmentation masks while generating a response. However, due to limitations induced by the grounding supervision, it often fails to precisely follow the human user's instructions (\eg, describing the image \emph{in detail}, answering the correct \emph{color}). \textbf{Middle:} Our approach unlocks and enhances the \emph{grounding ability implicitly learned by LMMs without explicit grounding supervision}, which leads to visually grounded responses while preserving the general vision-language conversation ability of LMMs. More examples are shown in Figure~\ref{fig:qual} in the supplementary material. \textbf{Right:} Previous methods train \textcolor{modelblue}{grounding LMMs} for visual grounding tasks at the cost of general visual question answering (VQA) performance. Our approach unlocks the implicit grounding ability in \textcolor{modelorange}{generalist LMMs} without further training and preserves their conversation ability.}
    \label{fig:teaser}
\end{strip}

\begin{abstract}
Current large multimodal models (LMMs) face challenges in visual grounding, which requires the model to relate language components to visual entities. Contrary to common practice that fine-tunes LMMs with additional grounding supervision, we find that grounding ability can be implicitly learned by LMMs to some extent without explicit grounding supervision that sacrifices general conversation ability. To unlock this grounding ability, we first introduce a training-free strategy ``\ourmethod,'' which analyzes the attention within an off-the-shelf LMM to provide a point prompt to a segmentation model (e.g., SAM) and perform pixel-level segmentation. This strategy instantly enables visual grounding for existing LMMs while keeping their original conversation ability intact. Second, motivated by vision-language alignment and localized features embedded in diffusion models, we propose \ourmodel---a LLaVA-like LMM that utilizes a diffusion-based visual encoder instead of the standard CLIP visual encoder. This design enhances the implicit grounding ability without changing the training data. Without being constrained by the biases and limited scale of grounding-specific supervision data, our approach enables strong visual grounding while preserving general conversation capabilities. We achieve competitive performance on both grounding-specific and general visual question answering benchmarks, compared with grounding LMMs and generalist LMMs, respectively. Notably, we achieve a 46.4 grounding mask recall on grounded conversation generation, outperforming the extensively supervised model GLaMM.
Project page: {\url{https://GroundLMM-ICCV.github.io}}.
\end{abstract}

\section{Introduction}
Large multimodal models (LMMs)~\citep{liu2023visual, zhu2024minigpt, dai2023instructblip} have brought a new opportunity to solve vision-language tasks in a general-purpose manner, which are typically built by connecting a visual encoder and a large language model (LLM) and fine-tuned by visual instructions. Currently, one major challenge faced by LMMs is \emph{visual grounding}---the key ability of relating language components (\eg, noun phrases) to visual entities (\eg, objects) in a given image~\citep{yu2016modeling, krishna2017visual}. With the grounding ability, LMMs can overcome the constraint of text-only responses and address more real-world vision-language tasks, such as robotics~\citep{gao2024physically, driess2023palm, szot2025grounding}.

To equip LMMs with the grounding ability, a common belief is that \emph{additional supervision for grounding} is necessary, and corresponding architectural changes must be introduced. For example, recent efforts extend the output modality from pure text to points~\citep{xu2024pixel, deitke2024molmo}, bounding boxes~\citep{chen2023shikra, peng2024grounding}, or segmentation masks~\citep{lai2024lisa, rasheed2024glamm}, by 1) attaching additional modules to the vanilla LMM architecture and 2) fine-tuning the LMM with grounding supervision. The grounding supervision originates from either re-purposing existing datasets with human-labeled object annotations or automatically annotating images using other models.

However, such \emph{reliance on strong supervision} introduces more undesired constraints: 1) \emph{Limited training data}: Image datasets with high-quality object-level annotations (at most millions of images~\citep{shao2019objects365, kuznetsova2020open}) rely on pre-defined categories and are significantly smaller than those with coarse image-text pairs (up to billions~\citep{schuhmann2022laion}), so re-purposing such object-level annotations only results in visual instruction data with limited diversity and quantity. Meanwhile, if the object-level annotations are produced by automated models, such annotations are noisier and less reliable than human-labeled ones~\citep{rasheed2024glamm}. 2) \emph{Supervision bias}: Changing the data focus to grounding tasks can lead to catastrophic forgetting~\citep{french1999catastrophic} and hurt LMMs' general conversation capabilities. Furthermore, whether the grounding data are manually annotated~\citep{lin2014microsoft} or pseudo-labeled by other models~\citep{rasheed2024glamm}, they are biased by the annotators' or models' knowledge and may fail to align with general human preferences, as these fine-grained annotations can vary significantly among different annotators or models. 3) \emph{Generalizability}: The grounding supervision is constrained within the visual concepts from either existing datasets or other models, which contradicts the ultimate goal of developing a general-purpose assistant to solve open-world problems~\citep{bendale2015towards}. Consequently, the resulting LMMs may \emph{be biased by the limited grounding supervision data, generalize poorly to novel visual concepts and domains, and lose general conversation abilities.}
Figure~\ref{fig:teaser} and Figure~\ref{fig:qual} in the supplementary material show examples of these limitations.

To avoid such limitations, the question worth rethinking then arises: \emph{Is there an approach to grounding LMMs other than strong supervision?} In fact, in this work, we reveal a critical yet previously overlooked fact: LMMs have inherently obtained some grounding ability through weakly supervised visual instruction tuning. In other words, \emph{the grounding ability can be learned to some extent implicitly by LMMs without grounding supervision.} Echoing prior observations of traditional convolutional neural networks~\citep{zhou2015object, zhou2016learning}, we find that LMMs learn to detect visual entities and relate them with the language implicitly, during the progress of vision-language learning at the image level.

We therefore propose a simple yet effective ``\ourmethod'' strategy in a training-free manner to \emph{unlock this implicit grounding ability in the form of pixel-level segmentation masks}, while maintaining the general conversation ability of LMMs. Intuitively, the attention mechanism~\citep{vaswani2017attention} in LMMs reveals \emph{where the LMM is looking at}, and thus provides clues for visual grounding. We start with a base LMM trained without grounding supervision (\eg, LLaVA~\citep{liu2023visual}), and acquire its \emph{attention corresponding to the visual input}. Though the entire attention map may be noisy, we locate the point where the LMM is focused on during token generation, and use the point to prompt a segmentation model (\eg, SAM~\citep{kirillov2023segment}\footnote{SAM is only trained for class-agnostic segmentation, and it has no visual grounding ability for relating visual elements with language.}) for accurate pixel-level grounding. With this \ourmethod method, we enable LMMs to perform vision-language tasks that directly require the grounding capability (\eg, grounded conversation generation~\citep{rasheed2024glamm}). Remarkably, \ourmethod does not require explicit grounding supervision like prior work; in contrast, \emph{weak supervision} from standard visual instruction tuning data is sufficient to achieve performance comparable with or even higher than previous grounding-supervised models in certain scenarios.

As a general approach, \ourmethod can be readily integrated with most recent generalist LMMs~\citep{chen2024expanding, bai2025qwen2}, and instantly unlock their grounding ability. From the visual representation perspective, we further introduce a simple solution to \emph{enhance the implicit grounding ability of LMMs}. Previously, CLIP~\citep{radford2021learning} plays a dominant role as the visual encoder of LMMs, due to its vision-language feature alignment. However, CLIP is known to be weak in providing localized visual features~\citep{zhou2022extract, ghiasi2022scaling, li2022language}, as its pre-training simply aligns the global representations of image-text pairs. In contrast, diffusion models~\citep{ho2020denoising, rombach2022high} can provide representations that are more suitable for visual grounding, as their text-to-image generation capability enables \emph{both vision-language alignment and localized features}. Thus, we propose a diffusion-based LMM (\ourmodel), which augments the CLIP visual encoder of the LMM with a diffusion-based visual encoder, while being fine-tuned using the \emph{same data} as the original LMM. By integrating an implicit captioner~\cite{xu2023open} and learnable positional encodings, we produce improved diffused-based visual features. Compared with the original LMM, \ourmodel strengthens the grounding ability without sacrificing performance in general-purpose vision-language tasks.

Our extensive experiments demonstrate that LMMs' grounding capabilities can be obtained using only weak vision-language supervision. Our approach, requiring no additional grounding supervision, \emph{does not suffer from biases in the grounding supervision data, and generalizes better} (see Figure~\ref{fig:teaser}). Despite being trained on less data than prior grounding LMMs~\citep{lai2024lisa, rasheed2024glamm}, our approach achieves better or comparable performance on grounding-specific benchmarks (\eg, 46.4 mask recall on the challenging grounded conversation generation task), while adhering to a strong generalist model for visual question answering tasks. To summarize, our contributions are three-fold:

\begin{itemize}[leftmargin=*, noitemsep, nolistsep]
    \item Different from prior methods that rely on strong grounding supervision, we show the possibility of unlocking the grounding ability in general LMMs without such supervision and preserving their general conversation ability.
    \item We discover a simple yet effective approach, \ourmethod, to achieve pixel-level grounding for LMMs by inspecting attention in token generation and prompting a segmentation model, which requires no training or architectural changes and is applicable to various LMMs.
    \item We propose \ourmodel, which employs a visual encoder based on the diffusion model. \ourmodel offers stronger grounding capabilities than LLaVA, while maintaining general VQA performance.
\end{itemize}

\section{Related Work}
\noindent\textbf{Large multimodal models (LMMs).}
Pioneering work in LMMs, such as LLaVA~\citep{liu2023visual, sun2024aligning, liu2024improved, liu2024llavanext}, MiniGPT-4~\citep{zhu2024minigpt, chen2023minigpt}, and InstructBLIP~\citep{dai2023instructblip, li2023blip}, enables visual inputs for large language models (LLMs) via vision-language feature alignment~\citep{radford2021learning} and instruction tuning~\citep{wei2022finetuned}. To equip LMMs with the grounding ability, a series of methods have been proposed to produce model outputs of bounding boxes~\citep{peng2024grounding, chen2023shikra, wang2023visionllm, pi2023detgpt, you2024ferret, li2024covlm}, traces of points~\citep{xu2024pixel}, or segmentation masks~\citep{lai2024lisa, rasheed2024glamm, zhang2024groundhog, ren2024pixellm}, by adding region-specific tokens or decoders. These methods require further grounding supervision, so image datasets with fine-grained annotations~\citep{lin2014microsoft, yu2016modeling, zhou2017scene} are usually repurposed for the visual instruction tuning. Unlike these grounding LMMs that follow the supervised paradigm, our approach, \ourmethod, does not change the LMM architecture or require any grounding supervision data. For the first time, we unlock LMMs' implicit grounding ability without relying on explicit supervision.
A concurrent work F-LMM~\citep{wu2024f} exploits attention maps in frozen LMMs for visual grounding, but we differ in two key aspects: 1) F-LMM follows the supervised paradigm, while our \ourmethod requires \emph{zero supervision}. For the first time, we reveal LMMs' implicit grounding capabilities without explicit supervision. 2) F-LMM examines existing LMMs without changing their visual encoding. In contrast, by analyzing visual encoders' grounding ability, we propose \ourmodel to further enhance implicit grounding.

\noindent\textbf{Diffusion models (DMs) as visual feature extractors.}
DMs~\citep{song2019generative, ho2020denoising, song2021score, karras2022elucidating, nichol2021improved, rombach2022high} have become a prevalent paradigm in visual generation, and intermediate features from DMs are explored for applications beyond generative tasks. For example, DDPM-Seg~\citep{baranchuk2022label}, ODISE~\citep{xu2023open}, and EmerDiff~\citep{namekata2024emerdiff} utilize DM features for various segmentation tasks. Features from DMs can also establish point- or pixel-level correspondences between images~\citep{tang2023emergent, luo2023diffusion, zhang2023tale, hedlin2023unsupervised}.
In this work, we show DMs can be utilized for learning a general-purpose LMM with strong grounding capabilities.

\section{Approach}
\label{sec:method}
In this section, we first introduce the common architecture design of LMMs (Section~\ref{sec:method-prelim}). We then discuss \ourmethod, which transforms the implicitly learned grounding ability into segmentation masks without training (Section~\ref{sec:method-aas}). Based on the standard LMM and \ourmethod, we propose \ourmodel, to further enhance the grounding ability without additional data (Section~\ref{sec:method-dlmm}). We include implementation details of our approach in the supplementary material.

\begin{figure*}[t]
    \centering
    \vspace{-4mm}
    \includegraphics[width=\textwidth]{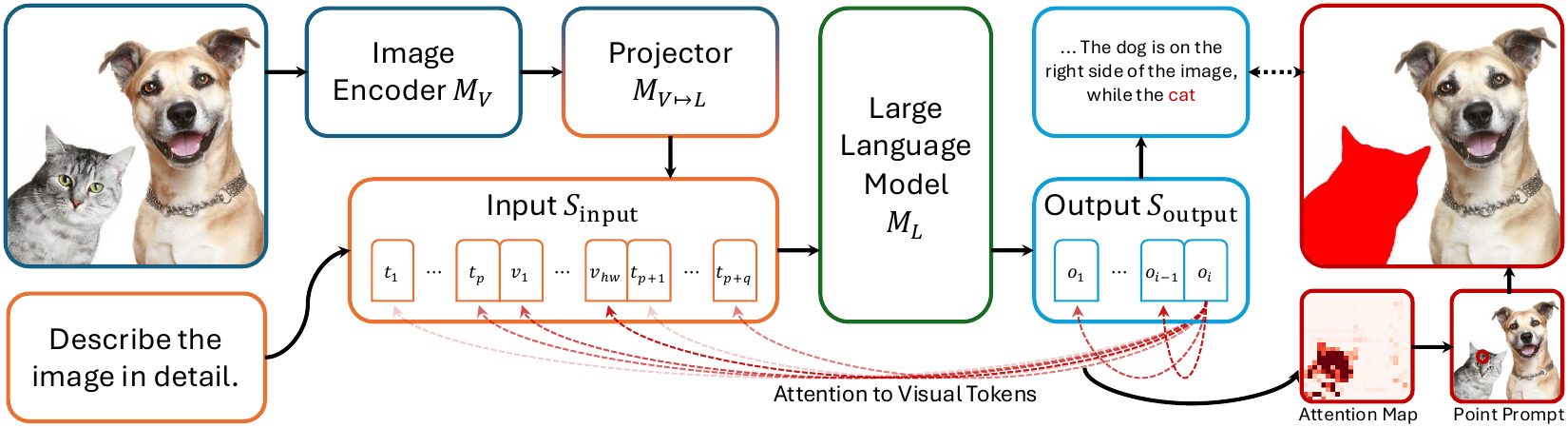}
    \caption{\textbf{Meta-architecture of LMMs and the \ourmethod strategy.} In a standard LMM, an image encoder $M_V$ extracts visual features from an input image, and the features are transformed into visual tokens by a lightweight projector $M_{V\mapsto L}$. A large language model $M_L$ generates outputs in an auto-regressive manner. When generating a new token (\eg, ``cat'') which requires grounding, we capture the \emph{attention} between the new token and the input visual tokens. Then a segmentation model (\eg, SAM~\citep{kirillov2023segment}) is prompted by the point with the highest normalized attention value to produce a \emph{segmentation mask} (\eg, cat in the image).}
    \label{fig:attention}
\end{figure*}

\subsection{Preliminary: Meta-Architecture of Large Multimodal Models}
\label{sec:method-prelim}

Most LMMs~\citep{liu2023visual, zhu2024minigpt, dai2023instructblip} share a common meta-architecture which consists of a \underline{v}isual encoder $M_V$, a \underline{v}ision-to-\underline{l}anguage feature projector $M_{V\mapsto L}$, and a large \underline{l}anguage model (LLM) $M_L$, as illustrated in Figure~\ref{fig:attention}. Given an image $I$ with resolution $H\times W$, the visual encoder $M_V$ (\eg, CLIP~\citep{radford2021learning}) is employed to extract visual features $V=M_V(I)\in\mathbb{R}^{h\times w\times c_V}$, where $h\times w$ represents the feature map size, and $c_V$ is the visual feature dimension. Then, the visual feature map is viewed as a sequence of $hw$ elements, and element-wise projected into the language feature space by the projector $M_{V\mapsto L}$. The projector can be implemented as a learnable multilayer perceptron (MLP). The $k$-th projected visual token is computed as $v_k=M_{V\mapsto L}(V_k)\in\mathbb{R}^{c_L}$, where $c_L$ is the feature dimension in the LLM. The visual tokens, concatenated with other language tokens, form the input sequence $S_\text{input}$:
\begin{equation}
S_\text{input}=\{t_1,\dots,t_p,v_1,\dots,v_{hw},t_{p+1},\dots,t_{p+q}\},
\end{equation}
where $\{v_1,\dots,v_{hw}\}$ are the $hw$ visual tokens projected from the visual feature map, $t_1,\dots,t_p$ are the $p$ language tokens before the visual tokens, and $\{t_{p+1},\dots,t_{p+q}\}$ are the $q$ language tokens after the visual tokens.

The LLM is usually a decoder-only transformer model that is capable of next-token prediction. Given the input sequence $S_\text{input}$, the output sequence $S_\text{output}=\{o_1,\dots,o_r\}$ is generated in an auto-regressive manner, where the $i$-th token is predicted as:
\begin{equation}
o_{i}=M_L(S_\text{input}, o_1,\dots, o_{i-1}).
\end{equation}
The generation terminates when the last predicted token $o_r$ is a special ``end-of-sequence'' token.

\begin{figure*}[t]
    \centering
    \vspace{-4mm}
    \includegraphics[width=\textwidth]{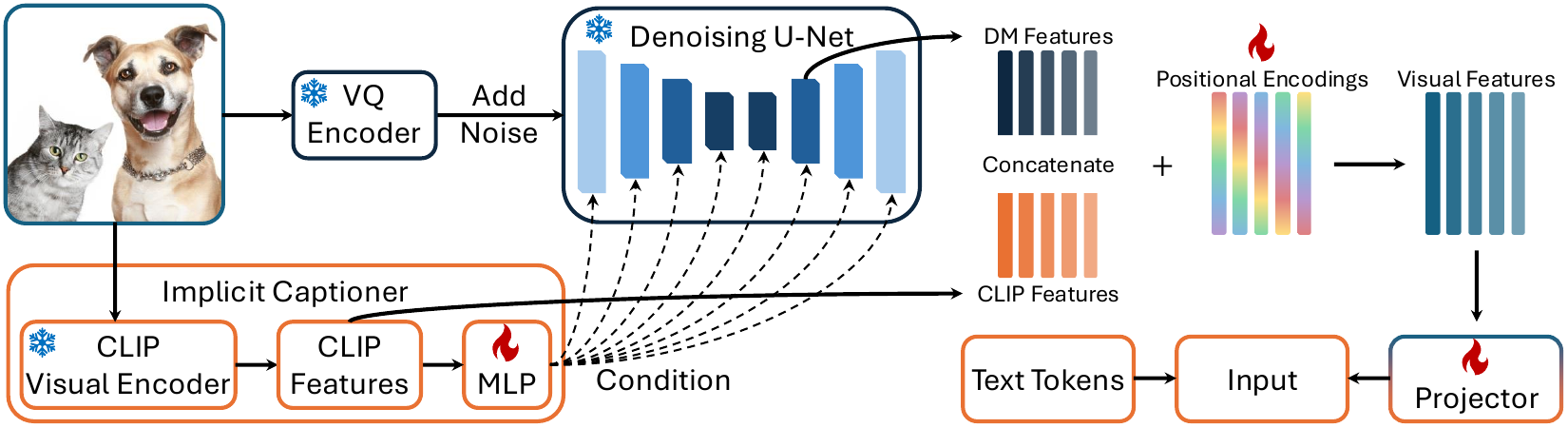}
    \caption{\textbf{Visual encoding in \ourmodel.} We perform one denoising step with a pre-trained diffusion model (DM)~\citep{ho2020denoising, rombach2022high}, and extract visual features from an intermediate block of the U-Net. The learnable implicit captioner~\citep{xu2023open} produces text-like conditioning and improves the visual features extraction in the U-Net. We combine both DM features and CLIP features, and add learnable positional encodings to them. The final visual features are projected into the language feature space via a learnable projector, and fed into the LLM along with other text tokens. The DM and CLIP visual encoder are pre-trained and frozen. This diffusion-based visual encoder does not significantly influence the overall efficiency, as the major computation happens in the LLM.}
    \label{fig:visual}
\end{figure*}

\subsection{\textsl{\textbf{Attend-and-Segment}}: Grounding LMMs Without Grounding Supervision}
\label{sec:method-aas}

Prior efforts towards grounding LMM attach a detection or segmentation module to the LMM architecture, and specialize the LMM training procedure with grounding supervision, \ie, visual instruction data augmented by object-level annotations, such that the LMM learns to predict connections between the text response and the image contents in the form of localized bounding boxes or segmentation masks. In contrast to these strongly supervised methods, we propose \ourmethod, a simple yet effective method for grounding LMMs \emph{without changing their architecture or requiring additional training}, as illustrated in Figure~\ref{fig:attention}. We investigate the attention maps inside the transformer-based language model when it generates tokens, and observe strong interpretablity associated with the attention maps. Intuitively, attention maps can provide information about \emph{where the model is looking at} when producing output language tokens.

Formally, we consider the input token sequence $S_\text{input}$ as detailed in Section~\ref{sec:method-prelim}. When predicting an output token $o_i$, we capture the raw attention maps $A_i^\text{raw}\in[0, 1]^{n_\text{layer}\times n_\text{head}\times(p+hw+q+i-1)}$ inside the transformer-based LLM $M_L$, where $n_\text{layer}$ is the number of layers in the LLM, $n_\text{head}$ is the number of heads per layer, and $p+hw+q+i-1$ is the number of tokens before the $i$-th output token $o_i$. We only use the attention maps associated with the $hw$ visual tokens and reduce the dimensions by averaging over $n_\text{layer}$ layers and $n_\text{head}$ heads per layer. This operation returns an attention matrix $A_i^\text{reduced}\in[0, 1]^{h\times w}$, with the same spatial dimension as the visual feature map.

Although being noisy, the attention between the output token and the visual tokens can provide interpretable grounding signals already. To further amplify the grounding signals and reduce the noise, we apply normalization across the whole output sequence:
\begin{equation}
\label{eq:attn}
A_i^\text{norm}=A_i^\text{reduced}-\frac{1}{r}\sum\nolimits_{j=1}^r A_j^\text{reduced},
\end{equation}
where $r$ is the output sequence length.

To provide pixel-level grounding, we derive a segmentation mask by upsampling the attention map and prompting a pre-trained segmentation model (\eg, SAM~\citep{kirillov2023segment}). For each token that requires grounding, we produce its corresponding binary mask by finding the point with the highest normalized attention and using its coordinate as a point prompt to the segmentation model. Thus, for elements of the output sequence, our \ourmethod method provides pixel-level grounding results. Notably, we use off-the-shelf segmentation models without modification, while fine-tuning is inevitable in prior pixel-level grounding LMMs~\citep{lai2024lisa, rasheed2024glamm}. In addition, the segmentation model does not need to be trained using language-guided grounding supervision. For example, SAM only learns to segment class-agnostic masks.

\subsection{\textsc{\textbf{DiffLMM}}: Enhanced Grounding With Diffusion-Based LMM}
\label{sec:method-dlmm}

Most LMMs employ CLIP~\citep{radford2021learning} as the visual encoder because it has been pre-trained to align vision and language representations, but CLIP is known to be sub-optimal in tasks that require precise localization (\eg, object detection, image segmentation)~\citep{zhou2022extract, ghiasi2022scaling, li2022language}. To enhance the grounding ability of LMMs, a direct choice may be replacing CLIP with better localized pure-vision backbones such as DINO~\citep{caron2021emerging, oquab2024dinov}. However, the lack of alignment with language representations can hurt vision-language capabilities and ultimately affect the visual grounding performance~\citep{jiang2023clip, tong2024eyes}.

Compared with vision-language models with image-level alignment (\eg, CLIP) and pure-vision models (\eg, DINO), visual representations from diffusion models (DMs) strike a better balance: 1) DMs learn to generate high-fidelity images, for which fine-grained and well-localized visual features are necessary. Consequently, they are better than CLIP in localization. 2) DMs are trained to perform text-to-image generation, and in this procedure, they acquire alignment with language instructions, which is lacking in pure-vision models. Therefore, we propose a diffusion-based LMM (\ourmodel, illustrated in Figure~\ref{fig:visual}), which strengthens the visual encoder with a pre-trained DM to improve the visual grounding ability and preserve the performance on general vision-language tasks.

To extract visual features for a given input image $I$, we simulate one denoising step in the diffusion process. The image is tokenized by a vector quantized (VQ) encoder, added with a random noise, and fed into the U-Net model of a DM~\citep{ho2020denoising, rombach2022high}. We extract the visual feature map from the second upsampling block in the U-Net, which best preserves semantic visual representations~\citep{tang2023emergent}. Text conditioning can enhance the visual feature extraction in the DM, but in practice, the image caption is usually unavailable. Therefore, we employ the implicit captioning mechanism~\citep{xu2023open}, which simulates the text conditioning with a CLIP visual encoder. Specifically, CLIP visual features are extracted as $V_\text{CLIP}=M_\text{CLIP}(I)$, projected by a multilayer perceptron (MLP) $M_{\text{CLIP}\mapsto\text{SD}}$, and injected into the U-Net via cross-attention modules. We denote the visual features of the DM as $V_\text{SD}=M_\text{SD}(I,M_{\text{CLIP}\mapsto\text{SD}}(V_\text{CLIP}))$. Finally, the visual feature map $V$ is composed by concatenating both DM features and CLIP features (note that we can reuse the CLIP features without additional overhead), and adding a set of learnable positional encodings $PE$~\citep{vaswani2017attention} to explicitly strengthen localization awareness in the features:
\begin{equation}
V=\text{concat}(V_\text{SD},V_\text{CLIP})+PE\in\mathbb{R}^{h\times w\times c_V}.
\end{equation}

For efficient training and preventing overfitting, we freeze pre-trained parameters in the CLIP visual encoder and the DM. Only the MLP in the implicit captioner, the positional encodings, and the vision-language feature projector are learnable in the visual encoder of \ourmodel. Since the computation is dominated by the LLM component in \ourmodel, integrating diffusion models in \ourmodel does not significantly affect efficiency. We only observe a marginal increase in the training and inference time ($<5\%$).

\begin{table*}[t]
    \centering
    \vspace{-4mm}
    % \resizebox{\textwidth}{!}{%
    \begin{tabular}{l|c c c|c c c}
        \toprule
        \multirow{2}{*}{Model} & \multicolumn{3}{c|}{GCG} & \multicolumn{3}{c}{VQA} \\
        & METEOR & mIoU & \textbf{Mask Recall} & VQAv2 & MMBench & MMStar \\
        \midrule
        \multicolumn{7}{l}{\cellcolor{cellgray}\textit{LMMs supervised by grounding tasks}} \\
        Kosmos-2$^\dagger$~\citep{peng2024grounding} & 15.8 & 56.8 & 29.0 & 45.6 & 59.2 & 24.9 \\
        LISA$^\dagger$$^\ddagger$~\citep{lai2024lisa} & 12.9 & 61.7 & 35.5 & 0.0 & 0.4 & 0.0 \\
        GLaMM$^\ddagger$~\citep{rasheed2024glamm} & 15.8 & 65.6 & 40.8 & 24.4 & 36.8 & 12.8 \\
        \midrule
        \multicolumn{7}{l}{\cellcolor{cellgray}\textit{LMMs not supervised by grounding tasks}} \\
        LLaVA-1.5~\citep{liu2024improved} + \textsl{a\&s} (Ours) & \bf 18.2 & 59.7 & 43.5 & 78.5 & 64.3 & 30.3 \\
        Qwen2.5-VL~\citep{bai2025qwen2} + \textsl{a\&s} (Ours) & 17.1 & 59.0 & 37.7 & \bf 84.0 & 83.5 & \bf 63.9 \\
        InternVL-2.5~\citep{chen2024expanding} + \textsl{a\&s} (Ours) & 18.0 & \bf 67.7 & 45.8 & 83.0 & \bf 84.6 & 62.8 \\
        \bf\ourmodel + \textsl{a\&s} (Ours) & \bf 18.2 & 63.3 & \bf 46.4 & 78.3 & 66.2 & 30.5 \\
        \bottomrule
    \end{tabular}%}

    \small $^\dagger$: GCG results are reported by GLaMM~\citep{rasheed2024glamm}. \small $^\ddagger$: VQAv2 and MMBench results are reported by F-LMM~\citep{wu2024f}.
    \caption{\textbf{Results on grounded conversation generation (GCG) and general visual question answering (VQA).} Even without grounding supervision, our \ourmethod (\textsl{a\&s} in the table) unlocks the implicitly learned grounding ability in LLaVA-1.5~\citep{liu2024improved} and other LMMs, \emph{outperforming all grounding-specific models} on GCG. \ourmodel further enhances the visual grounding ability and leads to better Mask Recall. As a general approach, \ourmethod can be applied on different LMMs~\citep{li2024llavanext-strong, tong2024cambrian}. Furthermore, while previous grounding-specific LMMs suffer from catastrophic forgetting and show degraded VQA performance, our approach well preserves the general conversation ability.}
    \label{tab:gcg_vqa}
\end{table*}

\section{Experiments}
\label{sec:expr}
In this section, we first present the results of applying our approach in both grounded conversation generation (Sections~\ref{sec:expr-gcg}) and general conversation tasks (Section~\ref{sec:expr-vqa}). We then provide the evaluation on two other representative visual grounding tasks (Section~\ref{sec:expr-respng}). Finally, we include an ablation study of our module designs (Section~\ref{sec:expr-ablation}). Due to limited space, we include implementation details, qualitative results, and additional analysis of LMM attention maps in the supplementary material. It is worth noting that \ourmethod and \ourmodel are general approaches for LMMs, but considering computational limitations, we focus on grounding and improving LMMs with 7B or 8B parameters~\citep{vicuna2023, meta2024llama}.

\subsection{Grounded Conversation Generation}
\label{sec:expr-gcg}

For evaluating the visual grounding ability in LMMs, we primarily focus on a \emph{comprehensive while challenging} visual grounding benchmark, grounded conversation generation (GCG)~\citep{rasheed2024glamm}, which requires the model not only to ground multiple visual entities in an image, but also to organize them into a localized description. Specifically, the GCG task requires the LMM to generate a detailed caption for a given image, in which noun phrases are related to their corresponding segmentation masks in the image.

Since GCG requires model abilities in both captioning and segmentation, three metrics are considered: 1) To measure the caption quality, the \emph{text-only metric}, METEOR~\citep{banerjee2005meteor}, compares the generated captions with the human-annotated reference captions. 2) To assess the segmentation quality, the \emph{mask-only metric}, mean intersection-over-union (mIoU), quantifies the similarity between ground-truth masks and their matched predicted masks. 3) The grounding mask recall~\citep{rasheed2024glamm} is an \emph{integrated metric} for region-specific grounding, which considers both the mask IoU and the textual similarities between the predictions and the ground truth. The integrated metric, grounding mask recall, is mainly considered when comparing different models.

Table~\ref{tab:gcg_vqa} compares our approach with previous grounding LMMs~\citep{peng2024grounding, lai2024lisa, rasheed2024glamm} on the test set of the Grand$_f$ dataset~\citep{rasheed2024glamm}. For this GCG task, our \ourmethod uses SAM~\citep{kirillov2023segment} as the segmentation model, and employs spaCy~\citep{spaCy} to parse model responses into noun phrases for grounding. \emph{Even without grounding supervision}, our \ourmethod leads to 43.5 mask recall for the original LLaVA-1.5~\citep{liu2024improved}, which is already \emph{higher than all the previous grounding LMMs}.
As a general approach, \ourmethod can be used in conjunction with recent LMMs such as Qwen2.5-VL~\cite{bai2025qwen2} and InternVL-2.5~\cite{chen2024expanding}, and unlock their implicitly learned visual grounding ability.
Compared to CLIP-based LMMs, \ourmodel provides better localized visual features and improves the grounding ability. Using our \ourmodel as the base LMM, we reach the highest 46.4 test recall. Our method achieves pixel grounding, but does not suffer from the supervision bias brought by grounding annotations, and thus better preserves the text-only conversation abilities, as shown by the higher METEOR scores.

\begin{table*}[ht]
    \centering
    \vspace{-4mm}
    % \resizebox{\textwidth}{!}{%
    \begin{tabular}{l|c c c|c c c|c c|c}
        \toprule
        \multirow{2}{*}{Model} & \multicolumn{3}{c|}{RefCOCO} & \multicolumn{3}{c|}{RefCOCO+} & \multicolumn{2}{c|}{RefCOCOg} & \multirow{2}{*}{\bf Avg.} \\
        & val & testA & testB & val & testA & testB & val & test & \\
        \midrule
        \multicolumn{10}{l}{\cellcolor{cellgray}\textit{Methods supervised on RES}} \\
        \color{gray}LISA~\citep{lai2024lisa} & \color{gray}74.9 & \color{gray}79.1 & \color{gray}72.3 & \color{gray}65.1 & \color{gray}70.8 & \color{gray}58.1 & \color{gray}67.9 & \color{gray}70.6 & \color{gray}69.9 \\
        \color{gray}GROUNDHOG~\citep{zhang2024groundhog} & \color{gray}78.5 & \color{gray}79.9 & \color{gray}75.7 & \color{gray}70.5 & \color{gray}75.0 & \color{gray}64.9 & \color{gray}74.1 & \color{gray}74.6 & \color{gray}74.2 \\
        \color{gray}GLaMM~\citep{rasheed2024glamm} & \color{gray}79.5 & \color{gray}83.2 & \color{gray}76.9 & \color{gray}72.6 & \color{gray}78.7 & \color{gray}64.6 & \color{gray}74.2 & \color{gray}74.9 & \color{gray}75.6 \\
        \color{gray}LLaVA-1.5 + F-LMM~\citep{wu2024f} & \color{gray}75.2 & \color{gray}79.1 & \color{gray}71.9 & \color{gray}63.7 & \color{gray}71.8 & \color{gray}54.7 & \color{gray}67.1 & \color{gray}68.1 & \color{gray}69.0 \\
        \midrule
        \multicolumn{10}{l}{\cellcolor{cellgray}\textit{Methods not supervised on RES}} \\
        Cropping~\citep{yu2023zero} & 22.7 & 21.1 & 23.1 & 24.1 & 22.4 & 23.9 & 28.7 & 27.5 & 24.2 \\
        Global-Local CLIP~\citep{yu2023zero} & 24.9 & 23.6 & 24.7 & 26.2 & 24.9 & 25.8 & 31.1 & 31.0 & 26.5 \\
        TAS~\citep{suo2023text} & 29.5 & 30.3 & 28.2 & 33.2 & 38.8 & 28.0 & 35.8 & 36.2 & 32.5 \\
        SAM-CLIP~\citep{ni2023ref} & 25.2 & 25.9 & 24.8 & 25.6 & 27.8 & 26.1 & 33.8 & 34.8 & 28.0 \\
        Ref-Diff~\citep{ni2023ref} & 35.2 & 37.4 & 34.5 & 35.6 & 38.7 & \bf 31.4 & \bf 38.6 & \bf 37.5 & 36.1 \\
        LLaVA-1.5~\citep{liu2024improved} + \textsl{a\&s} (Ours) & 41.0 & 51.6 & 29.5 & 33.0 & 43.7 & 24.4 & 32.4 & 31.6 & 35.9 \\
        Qwen2.5-VL~\citep{bai2025qwen2} + \textsl{a\&s} (Ours) & 37.3 & 44.5 & 32.1 & 33.7 & 41.2 & 28.8 & 28.8 & 28.3 & 34.3 \\
        InternVL-2.5~\citep{chen2024expanding} + \textsl{a\&s} (Ours) & 25.4 & 31.2 & 22.7 & 22.1 & 27.3 & 19.7 & 28.6 & 28.0 & 25.6 \\
        \ourmodel + \textsl{a\&s} (Ours) & \bf 46.8 & \bf 55.2 & \bf 38.1 & \bf 37.5 & \bf 46.0 & 30.0 & 34.4 & 34.8 & \bf 40.4 \\
        \bottomrule
    \end{tabular}%}
    \caption{\textbf{Results on referring expression segmentation (RES).} Without RES supervision, our approach obtains stronger ability to localize objects corresponding to given referring phrases, compared with prior methods that are also not trained on RES data. All models are evaluated by the cumulative intersection-over-union (cIoU) metric on RefCOCO(+/g)~\citep{yu2016modeling} datasets. For fair comparison, we mainly compare with methods that are not supervised on RES, and list supervised methods for reference.}
    \label{tab:res}
\end{table*}

\subsection{Visual Question Answering}
\label{sec:expr-vqa}
While enhancing the grounding ability of LMMs, we do not want LMMs to lose their general vision-language abilities. To assess such general abilities, we evaluate \ourmodel on a wide range of visual question answering (VQA) benchmarks, including VQAv2~\citep{goyal2017making}, MMBench~\citep{liu2024mmbench}, and MMStar~\citep{chen2024we}. More comparison between \ourmodel and LLaVA-1.5 on VQA benchmarks are included in Table~\ref{tab:vqa} in the supplementary material.

It is worth noting that previous grounding LMMs (\eg, LISA~\citep{lai2024lisa} and GLaMM~\citep{rasheed2024glamm}) are not usually evaluated on these general-purpose VQA benchmarks. For example, some questions are designed to examine object understanding in LMMs by asking questions like ``\texttt{Is there an [object] in the image?}'', but the queried object may not exist. However, we find that LISA and GLaMM almost always answers ``\texttt{Sure, it is [seg].}'' and provides an incorrect segmentation mask (see examples in Figure~\ref{fig:qual} in the supplementary material). Such loss of capabilities in answering general questions is due to the \emph{supervision bias}---these LMMs are fine-tuned for grounding tasks and they forget how to answer general visual questions without grounding. Therefore, grounding LMMs like GLaMM have extremely low scores on these benchmarks.

As shown in Table~\ref{tab:gcg_vqa}, compared with state-of-the-art LMMs of the same scale (fine-tuned from a 7B or 8B LLM), \ourmodel achieves performance on par with LLaVA-1.5, as \ourmodel is trained on the same data as LLaVA-1.5. Therefore, our diffusion-based \ourmodel improves fine-grained visual grounding ability while maintaining strong conversation ability as a generalist LMM.

\subsection{Referring Expression Segmentation and Panoptic Narrative Grounding}
\label{sec:expr-respng}

In addition to GCG, we provide additional results on two other widely investigated visual grounding tasks: \textbf{referring expression segmentation (RES)}~\citep{hu2016segmentation, yu2016modeling}, which segments a target object specified by a given referring expression, and \textbf{panoptic narrative grounding (PNG)}~\citep{gonzalez2021panoptic}, which grounds noun phrases in a given text description with panoptic segmentation masks. For better consistency with the two visual grounding tasks, we employ two segmentation models, Co-DETR~\citep{zong2023detrs} (for instance segmentation) and OpenSeeD~\citep{zhang2023simple} (for panoptic segmentation), in RES and PNG, respectively.

For fair comparison, we mainly consider baselines that are not supervised by RES/PNG data.
Notably, directly comparing our approach with prior supervised grounding LMMs is not exactly fair, due to the following reasons: 1) Our approach requires \emph{no grounding supervision}, while all prior grounding LMMs are extensively trained on such grounding tasks. 2) In both task settings, \emph{the text for grounding is set by an external input}, which is inconsistent with the generative nature of LMMs. Previous grounding LMMs can be supervised to adapt to such text specified by human users for better visual grounding results, while our approach does not have access to this opportunity. Therefore, in our evaluation, a conversation between a human user and an LMM is simulated to indirectly produce the attention maps and segmentation results.
Nevertheless, we achieve competitive performance on these two tasks, setting a new state of the art for methods that are not supervised on RES (+4.3 average cIoU) and PNG (+7.1 average recall), as shown in Tables~\ref{tab:res} and \ref{tab:png}.

\begin{table}[ht]
    \centering
    \vspace{-4mm}
    \begin{tabular}{l|c c c}
        \toprule
        Model & \bf All & Thing & Stuff \\
        \midrule
        \multicolumn{4}{l}{\cellcolor{cellgray}\textit{Methods supervised on PNG}} \\
        \color{gray}PixelLM$^\dagger$~\citep{ren2024pixellm} & \color{gray}43.1 & \color{gray}41.0 & \color{gray}47.9 \\
        \color{gray}GLaMM$^\dagger$~\citep{rasheed2024glamm} & \color{gray}55.8 & \color{gray}52.9 & \color{gray}62.3 \\
        \color{gray}GROUNDHOG~\citep{zhang2024groundhog} & \color{gray}66.8 & \color{gray}65.0 & \color{gray}69.4 \\
        \color{gray}LLaVA-1.5 + F-LMM~\citep{wu2024f} & \color{gray}64.8 & \color{gray}63.4 & \color{gray}68.2 \\
        \midrule
        \multicolumn{4}{l}{\cellcolor{cellgray}\textit{Methods not supervised on PNG}} \\
        DatasetDiffusion$^\ddagger$~\citep{nguyen2023dataset} & 23.5 & 16.0 & 33.8 \\
        DiffSeg$^\ddagger$~\citep{tian2024diffuse} & 24.1 & 17.7 & 33.0 \\
        DiffPNG~\citep{yang2024exploring} & 38.5 & 36.0 & 42.0 \\
        LLaVA-1.5~\citep{liu2024improved} + \textsl{a\&s} (Ours) & 42.8 & 35.2 & 53.6 \\
        Qwen2.5-VL~\citep{bai2025qwen2} + \textsl{a\&s} (Ours) & 40.1 & 30.8 & 53.3 \\
        InternVL-2.5~\citep{chen2024expanding} + \textsl{a\&s} (Ours) & 41.5 & 34.9 & 50.7 \\
        \ourmodel + \textsl{a\&s} (Ours) & \bf 45.6 & \bf 38.3 & \bf 55.8 \\
        \bottomrule
    \end{tabular}

    \small $^\dagger$: Reported by F-LMM~\citep{wu2024f}. $^\ddagger$: Reported by DiffPNG~\citep{yang2024exploring}.
    \caption{\textbf{Results on panoptic narrative grounding (PNG).} Our approach achieves the best recall among methods without grounding supervision. \ourmodel improves the original LLaVA-1.5 for visual grounding, which is consistent with our results on other tasks (Tables~\ref{tab:gcg_vqa} and \ref{tab:res}). The metric is average recall. For fair comparison, we mainly compare with methods that are not supervised on PNG, and list supervised methods for reference.}
    \label{tab:png}
\end{table}

\subsection{Ablation Study and Analysis}
\label{sec:expr-ablation}

\noindent\textbf{Processing attention maps.} Our \ourmethod applies normalization across the sequence of attention maps (Equation~\ref{eq:attn}), which significantly reduces noise in the maps (Figure~\ref{fig:attn-norm}).
From the attention map, we select the single point with the highest attention value to prompt SAM for the GCG task, instead of providing the entire map as a mask prompt.
Empirically, we find that attention maps are sparse, tending to focus on a few key points within objects rather than the entire objects, so point prompts are more effective. Quantitative comparisons are summarized in Table~\ref{tab:attn}. We use point-based prompts in all other experiments.

\noindent\textbf{Accuracy of point prompts.} The final performance metrics (\eg, mask recall in GCG and cIoU in RES) can be affected by both point prompts and segmentation quality. To separately assess the accuracy of point prompts, we consider the ratio of points that correctly fall into the true target region. We use PNG in this analysis because it covers both ``thing'' and ``stuff'' segmentation masks and the mask-text matching is fixed by the given narrative. As shown in Table~\ref{tab:point}, \ourmodel indeed improves the accuracy of point prompts.

\begin{table}[t]
    \centering
    \vspace{-4mm}
    % \resizebox{0.9\textwidth}{!}{%
    \begin{tabular}{c c|c c}
        \toprule
        Attention & SAM & \multicolumn{2}{c}{GCG} \\
        Norm & Prompt & mIoU & Mask Recall \\
        \midrule
        \cmark & Mask & 51.7 & 36.3 \\
        \xmark & Point & 59.4 & 43.9 \\
        \cellcolor{cellgray}\cmark & \cellcolor{cellgray}Point & \cellcolor{cellgray}\bf 63.3 & \cellcolor{cellgray}\bf 46.4 \\
        \bottomrule
    \end{tabular}%}
    \caption{\textbf{Ablation study on \ourmethod.} Normalizing attention maps across the entire sequence removes noisy patterns and improves grounding. Prompting SAM~\citep{kirillov2023segment} with a single point instead of a low-resolution mask is more effective. Our \ourmethod combines both techniques. The results are based on evaluating \ourmodel on the GCG task~\citep{rasheed2024glamm}.}
    \label{tab:attn}
\end{table}

\begin{table}[t]
    \centering
    % \resizebox{0.9\textwidth}{!}{%
    \begin{tabular}{l|c c c}
        \toprule
        \multirow{2}{*}{Model} & \multicolumn{3}{c}{Point Accuracy} \\
        & All & Thing & Stuff \\
        \midrule
        LLaVA-1.5~\citep{liu2024improved} & 52.92 & 44.33 & 64.98 \\
        \ourmodel & \bf 56.74 & \bf 48.74 & \bf 67.98 \\
        \bottomrule
    \end{tabular}%}
    \caption{\textbf{Accuracy of point prompts by different LMMs.} \ourmodel improves the implicit grounding ability of LLaVA-1.5 and its attention maps lead to more accurate point prompts for PNG.}
    \label{tab:point}
\end{table}

\begin{table}[t]
    \centering
    % \resizebox{0.8\textwidth}{!}{%
    \begin{tabular}{l|c c|c}
        \toprule
        Backbone & PE & IC & Mask Recall \\
        \midrule
        CLIP (LLaVA-1.5) & \multicolumn{2}{c|}{--} & 43.5 \\
        DINOv2 & \multicolumn{2}{c|}{--} & 41.9 \\
        DINOv2 + CLIP & \multicolumn{2}{c|}{--} & 44.1 \\
        \midrule
        SD-1.5 & & & 41.8 \\
        SD-1.5 & \cmark & & 42.0 \\
        SD-1.5 & \cmark & \cmark & 44.0 \\
        \cellcolor{cellgray}SD-1.5 + CLIP (\ourmodel) & \cellcolor{cellgray}\cmark & \cellcolor{cellgray}\cmark & \cellcolor{cellgray}\bf 46.4 \\
        \bottomrule
    \end{tabular}%}
    \caption{\textbf{Ablation study on \ourmodel.} Both positional encodings (PE) and the implicit captioner (IC) improve the grounding ability of \ourmodel.}
    \label{tab:visual}
\end{table}

\noindent\textbf{Visual encoding.} In \ourmodel (Figure~\ref{fig:visual}), we employ a few modules to enhance the visual feature extraction, including learnable \emph{positional encodings}~\citep{vaswani2017attention} and an \emph{implicit captioner}~\citep{xu2023open} that simulates text conditioning with CLIP visual features.
In Table~\ref{tab:visual}, we compare the impact of various visual encoders and the design choices of \ourmodel on the grounding ability, evaluated by GCG mask recall. Integrated with positional encodings the implicit captioner, \ourmodel achieves better grounding performance than trivially replacing or combining CLIP~\citep{radford2021learning} with a pure-vision encoder, DINOv2~\citep{oquab2024dinov}.

\section{Conclusion}
In this work, we reveal a previously overlooked yet critical fact that LMMs implicitly obtain grounding capabilities even if they are trained \emph{without} grounding supervision. We propose \ourmethod to unlock this implicit grounding ability and produce segmentation masks, and introduce \ourmodel to further enhance this grounding ability. Different from models extensively supervised by grounding data, our approach can easily adapt generalist LMMs for grounding without training and preserve their general conversation ability. Moreover, extensive evaluation results demonstrate strong performance on both grounding-specific and general vision-language benchmarks, even surpassing grounding LMMs trained with extensive supervision on the challenging grounded conversation generation task.

\paragraph{Acknowledgments.} 
This work was supported in part by NSF Grant 2106825, NIFA Award 2020-67021-32799, the Amazon-Illinois Center on AI for Interactive Conversational Experiences, the Toyota Research Institute, the IBM-Illinois Discovery Accelerator Institute, and Snap Inc. This work used computational resources, including the NCSA Delta and DeltaAI supercomputers through allocations CIS230012, CIS230013, CIS240133, and CIS240428 from the Advanced Cyberinfrastructure Coordination Ecosystem: Services \& Support (ACCESS) program, as well as the TACC Frontera supercomputer, Amazon Web Services (AWS), and OpenAI API through the National Artificial Intelligence Research Resource (NAIRR) Pilot.

{
    \small
    \bibliographystyle{ieeenat_fullname}
    \bibliography{main}
}

\clearpage
\appendix
\maketitlesupplementary
\setcounter{section}{0}
\setcounter{figure}{0}
\setcounter{table}{0}
\renewcommand{\thesection}{\Alph{section}}
\renewcommand{\thefigure}{\Alph{figure}}
\renewcommand{\thetable}{\Alph{table}}

\noindent In this supplementary material, we provide additional details of our implementation (Section~\ref{app:impl}) and visual grounding tasks (Section~\ref{app:grounding}), qualitative results (Section~\ref{app:qual}), evaluation on VQA benchmarks (Section~\ref{app:vqa}), and analysis of LMM attention maps (Section~\ref{app:attn}).

\section{Implementation Details of Our Approach}
\label{app:impl}
In this section, we provide the implementation details of this work to ensure reproducibility of our experiments.

\noindent\textbf{\textsl{Attend-and-Segment}.} We first collect the attention maps for the visual tokens, and aggregate the attention maps by averaging over all layers and heads. Then, we apply normalization across the output token sequence to remove noisy points and upsample the normalized attention map to the original image resolution. During mask generation, we find the coordinate where the normalized attention value is maximized, and use it as a prompt to the segmentation model (\eg, SAM~\citep{kirillov2023segment}) for producing the pixel-level segmentation map. In the grounded conversation generation task, we parse the model response into noun phrases using \texttt{spaCy}~\citep{spaCy}. If a noun phrase contains multiple tokens, we still use the point with the highest normalized attention value across all tokens within the phrase as the prompt.

\noindent\textbf{\ourmodel.} Our development of \ourmodel is based on the codebase and dataset of LLaVA-1.5~\citep{liu2024improved}. We employ the Stable Diffusion v1.5~\citep{rombach2022high} model as our visual backbone. In the denoising step, we add a random noise at the 100 timestep and extract features from the second upsampling block, following the practice of DIFT~\citep{tang2023emergent}. We also provide an ablation study on the choice of the noise level and feature block in Table~\ref{tab:diffusion}. In the implicit captioner~\citep{xu2023open}, we employ the visual encoder of CLIP-ViT-L-336px~\citep{radford2021learning}, the same CLIP model in the original LLaVA-1.5. The model is trained with LoRA~\citep{hu2022lora} and the same training recipe and same training data as LLaVA-1.5.

\begin{table}[ht]
    \centering
    \begin{tabular}{c|c|c}
        \toprule
        Noise Step & Feature Block & Pre-train Loss $\downarrow$ \\
        \midrule
        \cellcolor{cellgray}100 & \cellcolor{cellgray}2 & \bf 2.384 \\
        \midrule
        0 & \multirow{3}{*}{2} & 2.417 \\
        200 & & 2.395 \\
        300 & & 2.457 \\
        \midrule
        \multirow{3}{*}{100} & 1 & 2.400 \\
        & 3 & 2.465 \\
        & 4 & 2.625 \\
        \bottomrule
    \end{tabular}
    \caption{\textbf{Ablation study on diffusion feature extraction.} Adding a relatively small noise (at diffusion step 100 or 200) to the original image and extracting features from the second upsampling block in the diffusion U-Net lead to the best results in \ourmodel.}
    \label{tab:diffusion}
\end{table}

\section{Details of Referring Expression Segmentation and Panoptic Narrative Grounding}
\label{app:grounding}

\noindent\textbf{Referring expression segmentation (RES).} The RES task requires the model to segment a target object specified by a given referring expression. It is indirect for our method, \ourmethod, to provide a single attention map for the referring expression specified by a task input. Therefore, we design a simulation strategy to find the text-image correspondence in RES. Specifically, we formulate a one-round conversation between a human user and an LMM. For a referring expression \texttt{[expr]}, the human user asks the model to ``\texttt{Describe the [expr].}'' Then, the model is expected to generate a response that focuses on the target object. We find the first token of the object of the sentence (\ie, the token right after the verb ``is'' ``are'' or ``to be''), and extract its attention map. After that, we can apply \ourmethod to find the point with the highest attention value, and produce a grounding mask based on it. An illustrative example is shown as follows, in which we will use the token ``\texttt{\textbf{a}}'' for attention map extraction.

\texttt{USER: Describe the "middle player."}

\texttt{MODEL: The "middle player" in the image is \textbf{a} baseball player wearing a blue shirt and a baseball glove...}

We observe that SAM~\citep{kirillov2023segment} tends to produce overly fine-grained segmentation masks for object parts, and cannot deliver satisfactory results in the RES task which requires instance segmentation masks for entire objects. Since our approach is \emph{not limited to any specific segmentation model}, we can use the prompt point generated by \ourmethod to guide other models for segmentation. In this experiment, we use a Co-DETR~\citep{zong2023detrs} instance segmentation model to produce class-agnostic mask predictions for each image. To avoid data contamination, we exclude RefCOCO(+/g) validation/test images from the training set and retrain the Co-DETR model. After generating both the prompt point with \ourmethod and the set of candidate masks, we select the mask that contains the prompt point. If there are multiple masks containing the prompt point, we just use the mask with the highest confidence score predicted by Co-DETR.

\begin{table*}[t]
    \centering
    \begin{tabular}{l|c c c c c c c c c}
        \toprule
        Model & VQAv2 & GQA & VW & SQA & TQA & POPE & MM-B & LV-B & MM-S \\
        \midrule
        LLaVA-1.5~\citep{liu2024improved} & \bf 78.5 & 62.0 & \bf 50.0 & 66.8 & \bf 58.2 & \bf 85.9 & 64.3 & \bf 65.4 & 30.3 \\
        \bf\ourmodel (Ours) & 78.3 & \bf 62.1 & 48.1 & \bf 69.3 & 57.2 & 85.7 & \bf 66.2 & 63.7 & \bf 30.5 \\
        \bottomrule
    \end{tabular}
    \caption{\textbf{Visual Question Answering (VQA) results.} We evaluate \ourmodel on a wide range of benchmarks, including VQAv2~\citep{goyal2017making}, GQA~\citep{hudson2019gqa}, Vizwiz (VW)~\citep{gurari2018vizwiz}, ScienceQA-IMG (SQA)~\citep{lu2022learn}, TextVQA (TQA)~\citep{singh2019towards}, POPE~\citep{li2023evaluating}, MMBench (MM-B)~\citep{liu2024mmbench}, LLaVA-Bench (LV-B)~\citep{liu2023visual}, and MMStar (MM-S)~\citep{chen2024we}. Different from prior models, \ourmodel is built upon a diffusion-based visual encoder, which provides stronger grounding (Tables~\ref{tab:gcg_vqa}, \ref{tab:res}, and \ref{tab:png}) and preserves vision-language abilities in general tasks.}
    \label{tab:vqa}
\end{table*}

The results are summarized in Table~\ref{tab:res}. Without any training on RES or other visual grounding tasks, we achieve a remarkable performance of 40.4 average cIoU in RES. In particular, we outperform the previous zero-shot methods~\citep{yu2023zero, suo2023text, ni2023ref}. The results demonstrate strong visual grounding capabilities implicitly learned by LMMs.

\noindent\textbf{Panoptic narrative grounding (PNG).} The PNG task requires the model to ground each noun phrase in a given text description of a given image by providing a corresponding panoptic segmentation mask~\citep{kirillov2019panoptic} for each noun phrase. Different from the standard setup of LMMs, the text description of the image is provided by the task, rather than generated by the model itself. Therefore, we need to adapt our \ourmethod method for the PNG task. Specifically, we simulate a one-round conversation between a human user and an LMM. The human user asks the model to ``\texttt{Describe the image in detail.}'' Then, the model responds with the given text description. We extract attention maps from the model response part of this conversation and use the attention maps to guide the segmentation procedure.

Similar to RES, we use a panoptic segmentation model, OpenSeeD~\citep{zhang2023simple}, to provide high-quality panoptic segmentation masks for each image. For each noun phrase, we find all the associated token and their attention maps. Then, we locate the point with the highest attention value and select the mask (from all the candidate masks generated by OpenSeeD) that contains this point. Note that in panoptic segmentation, all masks are non-overlapping, so there is only one mask that contains this point of the highest attention value. This selected mask is predicted as the region corresponding to the noun phrase.

In Table~\ref{tab:png}, we compare the results of our approach with previous methods. Although our approach is applied on the PNG task in a unsupervised manner, it achieves competitive performance and even outperforms one prior grounding LMM, PixelLM. Additionally, our approach outperforms all previous PNG methods that are not supervised on PNG data~\citep{nguyen2023dataset, tian2024diffuse, yang2024exploring}.

\section{Evaluation on VQA Benchmarks}
\label{app:vqa}

In Table~\ref{tab:vqa}, we list a thorough evaluation of \ourmodel, and compare it with the baseline model LLaVA-1.5~\citep{liu2024improved}. Since they share the same training data, we observe similar performance between LLaVA-1.5 and \ourmodel, suggesting that the diffusion-based visual encoder in \ourmodel can preserve the general conversation capabilities.

\begin{figure*}[t]
    \centering
    \includegraphics[width=\textwidth]{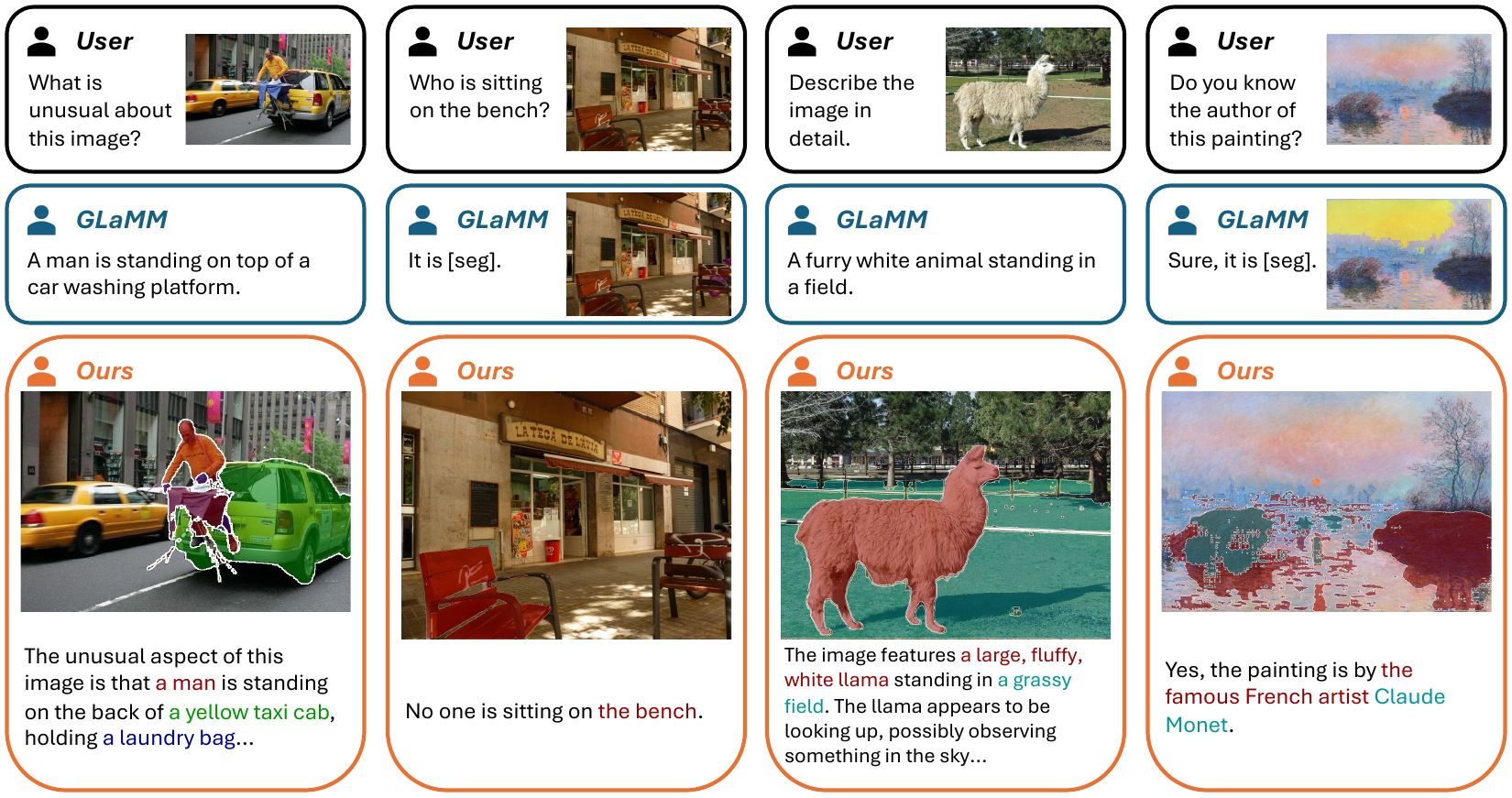}
    \caption{\textbf{Comparison of model responses to challenging visual questions.} 1) \emph{Unusual image contents}: The model is requested to analyze the unusual aspect of a given image. Compared with GLaMM, our approach provides a more detailed and accurate answer with grounding. 2) \emph{Adversarial questions}: The model is asked about something that does not exist in the image. GLaMM insists to segment the bike behind the bench in this example. 3) \emph{Rare visual concepts}: The image contains objects of less frequent categories. In this example, GLaMM does not recognize the llama but describes it in a general manner, while our approach provides a more accurate description. 4) \emph{Shifted image domain}: An image from a new domain is given to the model. Interestingly, our approach seems to be making the decision based on the texture and style in the painting. For visual clarity, we only show the beginning parts of our model responses if they are too long. These challenging examples demonstrates better \emph{generalizability of our approach}.}
    \label{fig:qual}
\end{figure*}

\section{Qualitative Results}
\label{app:qual}

In Figure~\ref{fig:qual} we present qualitative results of \ourmodel + \ourmethod for more challenging visual questions that are different from the training data, in comparison with GLaMM~\citep{rasheed2024glamm}. First, when the questions are not formulated as usual, GLaMM tends to interpret these questions as image captioning or referring expression segmentation tasks, while \ourmodel can still follow the user's instructions and accurately answer the questions. Meanwhile, \ourmethod provides well-grounded responses that connects text phrases and visual entities. Furthermore, our approach shows \emph{better generalizability to unfamiliar} question types, visual concepts, and image domains.

\begin{figure*}[t]
    \centering
    \includegraphics[width=0.8\textwidth]{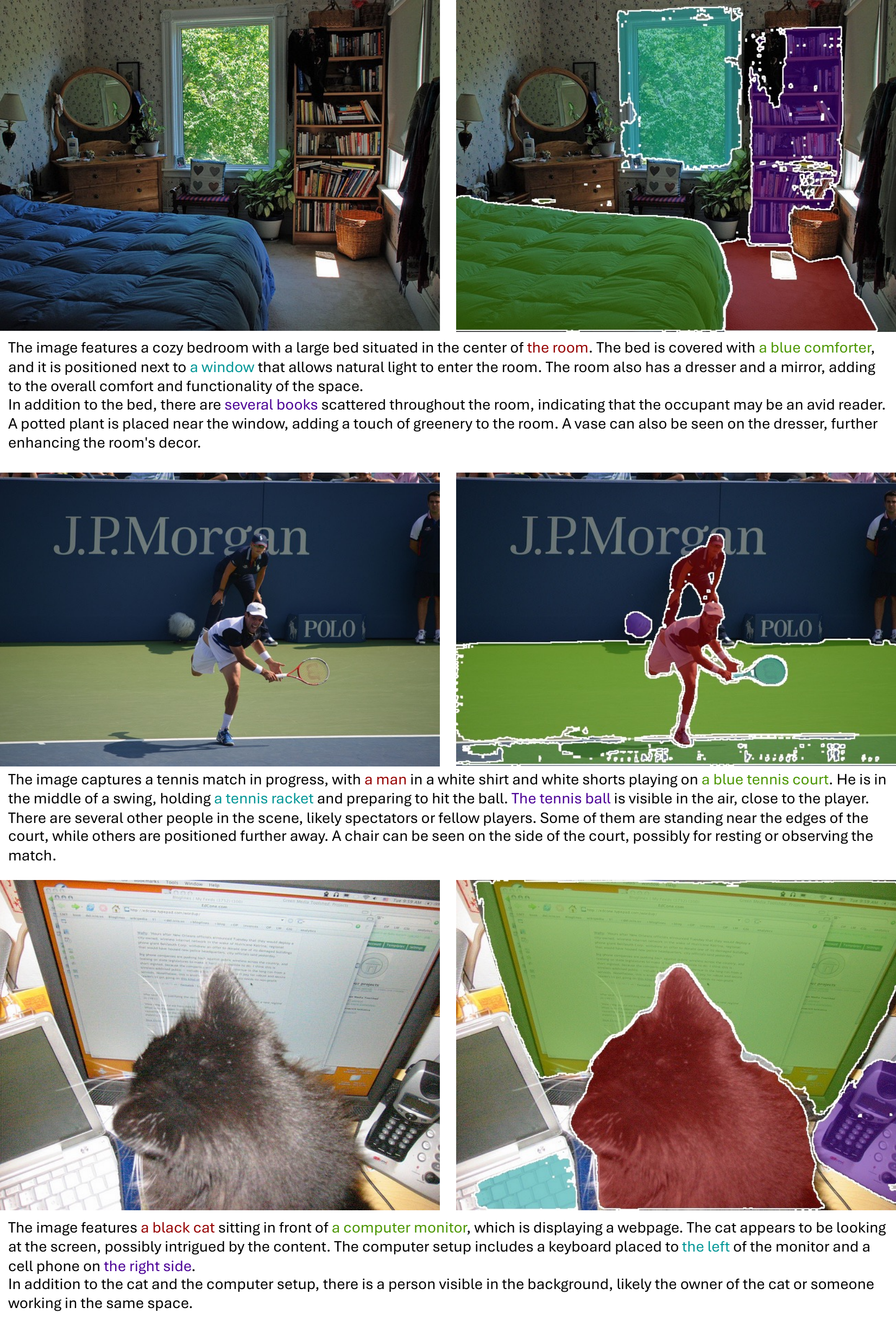}
    \caption{\textbf{Qualitative results for grounded conversation generation.} For visual clarity, we only display the best four non-overlapping segmentation masks per image.}
    \label{fig:qual-gcg}
\end{figure*}

We present additional qualitative results for the grounded conversation generation task in Figure~\ref{fig:qual-gcg}. The \ourmodel model is asked to ``\texttt{Describe the image in detail.}'' Then we use \ourmethod to produce visual grounding. Overall, our approach can provide accurate segmentation masks, but may also suffer from common issues of LMMs (\eg, object hallucination~\citep{li2023evaluating, sun2024aligning}).

\section{Additional Analysis of Attention Maps}
\label{app:attn}
In \ourmethod, we aggregate the attention values between each generated token and the visual tokens into a 2D map. In this section, we provide a more in-depth analysis of the attention maps. For visualization, we use the same ``cat and dog'' image (Figure~\ref{fig:teaser}) as an example in the following analysis; we have similar observations on other images as well.

\begin{table}[ht]
    \centering
    \begin{tabular}{c c|c c c c|c}
        \toprule
        & & \multicolumn{4}{c|}{Head Index} & \multirow{2}{*}{avg.$\pm$std.} \\
        & & 1 & 9 & 17 & 25 \\
        \midrule
        & 1 & 19.2 & 13.6 & 19.9 & 11.9 & 16.2$\pm$3.5 \\
        Layer & 9 & 7.5 & 25.1 & 9.0 & 28.2 & 17.5$\pm$9.3 \\
        Index & 17 & 26.2 & 4.3 & 19.7 & 27.1 & 19.3$\pm$9.1 \\
        & 25 & 5.9 & 34.3 & 27.0 & 15.7 & 20.7$\pm$10.8 \\
        \midrule
        \multicolumn{6}{c|}{Overall} & 18.4$\pm$8.8 \\
        \bottomrule
    \end{tabular}
    \caption{\textbf{Evaluation of attention maps from individual head/layer combinations.} Applying \ourmethod on the attention maps extracted from individual heads and layers results in worse and less stable grounding mask recall in GCG, as compared with applying \ourmethod on the mean attention maps aggregated over all heads and layers, which achieves 46.4 mask recall (Table~\ref{tab:gcg_vqa}).}
    \label{tab:layer-head}
\end{table}

\begin{figure*}
    \centering
    \includegraphics[width=\textwidth]{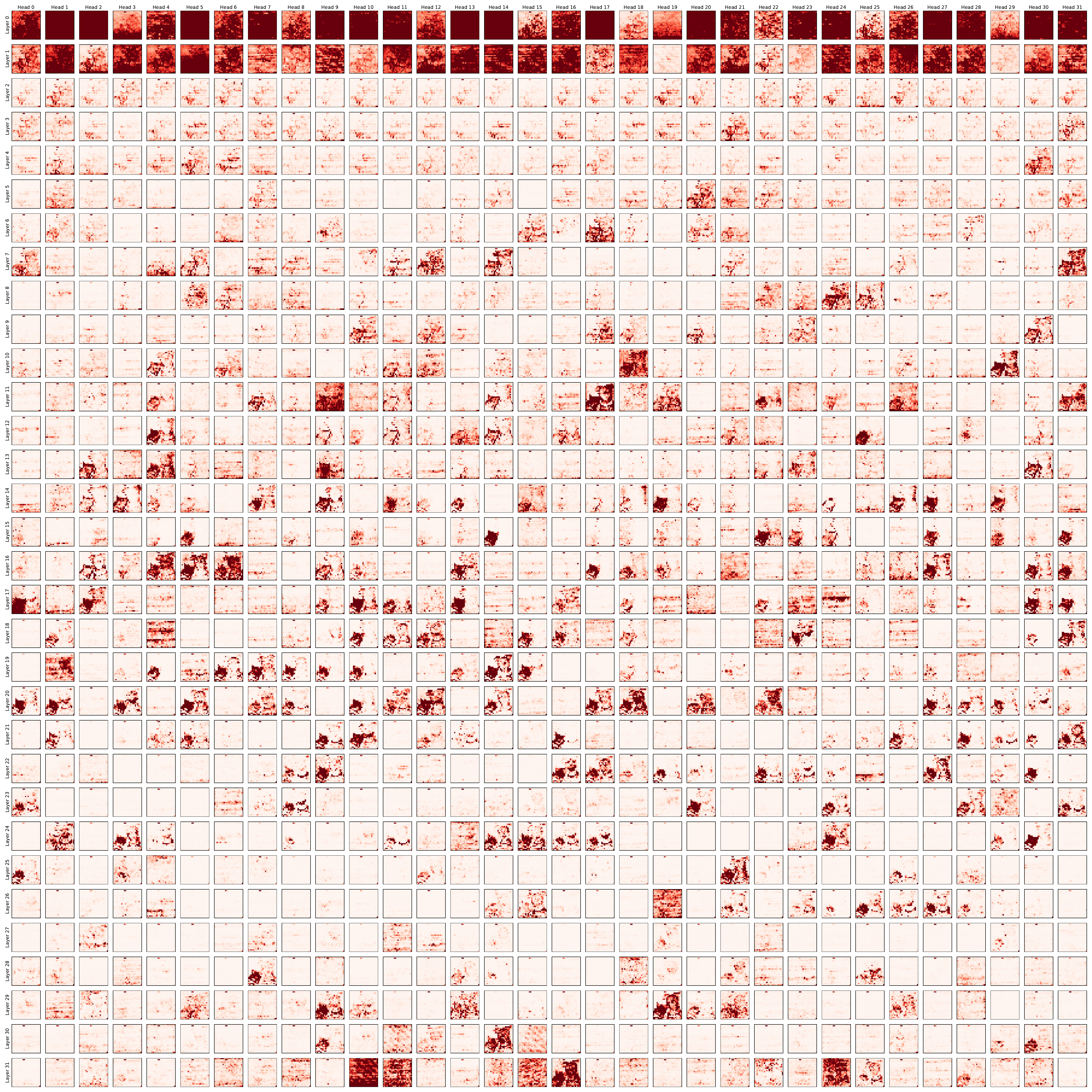}
    \caption{\textbf{Attention between the visual tokens and the generated token ``cat.''} We observe certain heads in the intermediate layers produce more localized attention maps with respect to the ``cat'' object in the image (\eg, Head 14 of Layer 15). It remains challenging to directly relate individual heads to visual concepts when grounding annotations are not available, so \ourmethod directly aggregates attention maps from all layers and heads by averaging them.}
    \label{fig:attn-head}
\end{figure*}

\noindent\textbf{Attention in each head and layer.} Instead of averaging the attention values over $n_\text{layer}$ layers and $n_\text{head}$ heads per layer in the LLM, we first inspect the individual attention values in each head and layer. Figure~\ref{fig:attn-head} visualizes the attention between one generated token ``cat'' and the input visual tokens. Consistent with some recent observations~\citep{wu2024retrieval}, a few heads in the intermediate layers show stronger activation with respect to the visual object in the image. Also, attention maps in intermediate layers are more localized. However, it is infeasible to build direct connections between attention heads and visual concepts, given the absence of grounding annotations. 

Table~\ref{tab:layer-head} summarizes an empirical study that demonstrates the grounding results of using the attention from one single head of one single layer. Compared with averaging over all heads and layers, individual heads and layers lead to to significantly worse and noisier results. Therefore, we aggregate the attention maps across all heads and layers by averaging, which also simplifies the algorithm of \ourmethod in our setting without grounding supervision.

\begin{figure*}
    \centering
    \includegraphics[width=\textwidth]{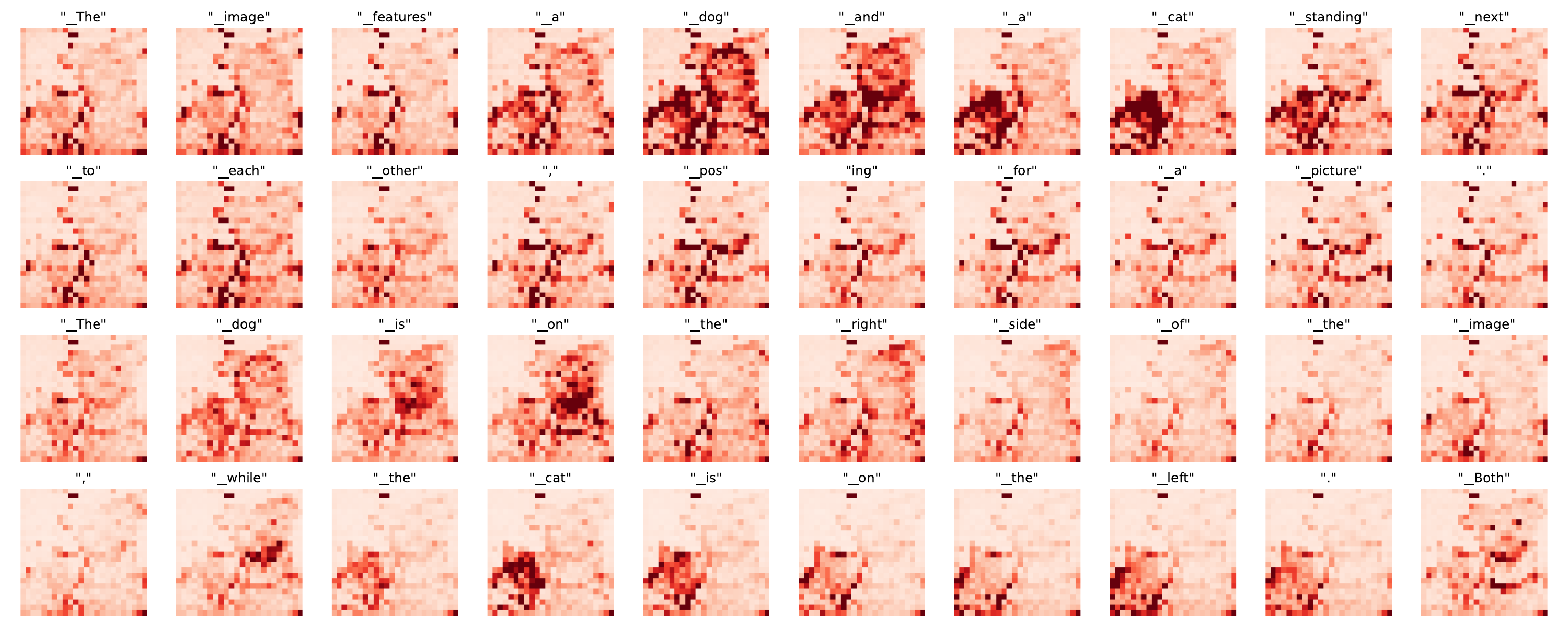}

    \vspace{4mm}

    \includegraphics[width=\textwidth]{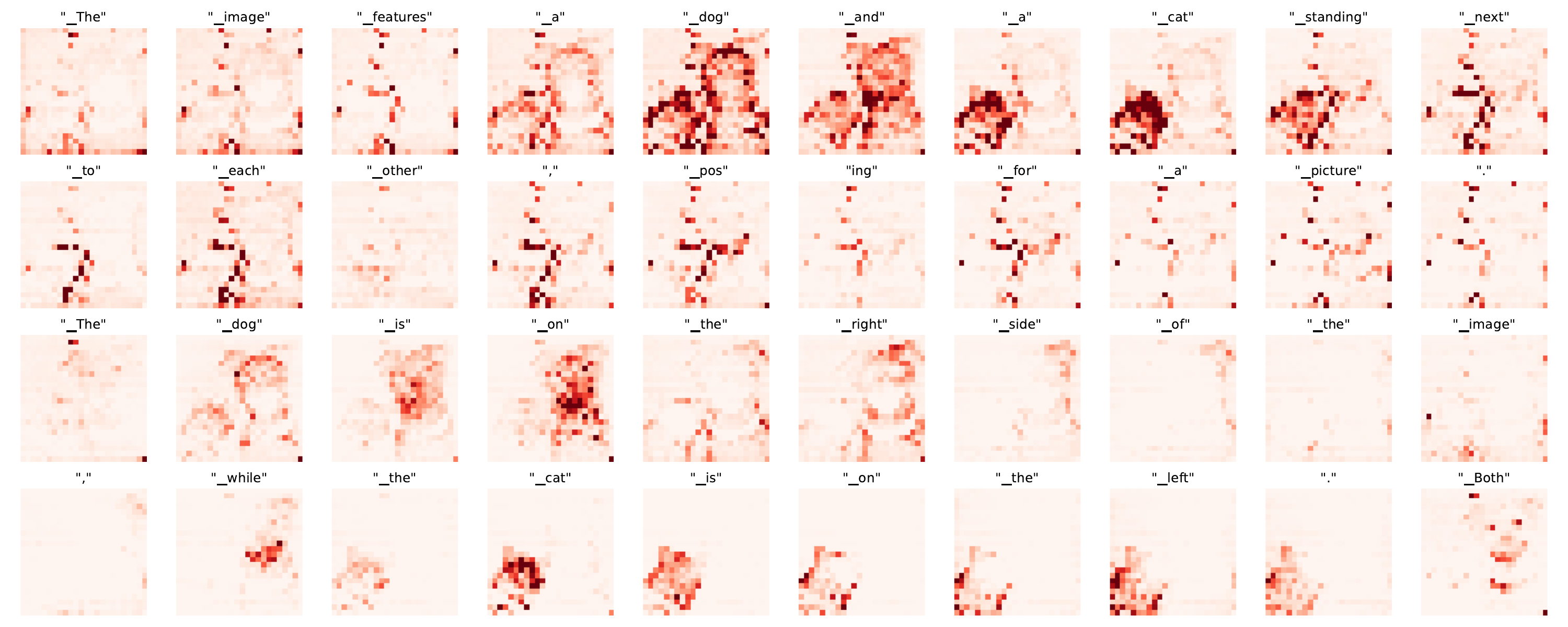}
    \caption{\textbf{Attention maps before and after the normalization.} Top: Before the normalization, a few uninformative visual tokens in the background (\eg, top-center tokens above the dog's head) receive more attention, which is consistent with the observation of ``register tokens''~\citep{darcet2024vision}. Such patterns are stable across the output sequence. Bottom: To remove such artifacts in the attention maps, we subtract the mean attention map (Section~\ref{sec:method-aas}). After the normalization, the attention maps show clearer localization, and are suitable for pixel-level grounding. In addition to noun phrases, other parts of the text response demonstrate meaningful visual correspondence (\eg, ``next to each other'' corresponds to the space between the two animals).}
    \label{fig:attn-norm}
\end{figure*}

\noindent\textbf{Attention normalization.} After reducing the attention maps into one 2D map for each generated token, we observe some noisy patterns in the attention maps (Figure~\ref{fig:attn-norm}-top). Some seemingly uninformative visual tokens (usually in the background) attract more attention from the generated token than other visual tokens. A recent work~\citep{darcet2024vision} shows similar observations, and explains that such less informative tokens are ``repurposed for internal computations.'' To remove such artifacts, they propose to provide additional tokens to the vision transformer as registers.

However, in our setting, we cannot retrain the visual backbone or the language model due to limited data and computation. Instead, we simply normalize the attention maps by subtracting the mean attention map averaged over the output sequence (Section~\ref{sec:method-aas}). Although the noisy attention patterns exist, we observe that these patterns are relatively stable (Figure~\ref{fig:attn-norm}-top), so the mean attention map, aggregated over the output sequence, can capture the undesired attention patterns and allow us to remove them.

After the attention normalization, we observe clearer patterns (Figure~\ref{fig:attn-norm}-bottom) which leads to accurate pixel grounding. Quantitatively, attention normalization improves the GCG mask recall from 43.9 to 46.4 (Table~\ref{tab:attn}). In addition to noun phrases, other words reveal relations or comparisons between visual entities, and could be helpful for more vision-language tasks. We leave this investigation for future research.

\begin{figure*}
    \centering
    \includegraphics[width=0.8\textwidth]{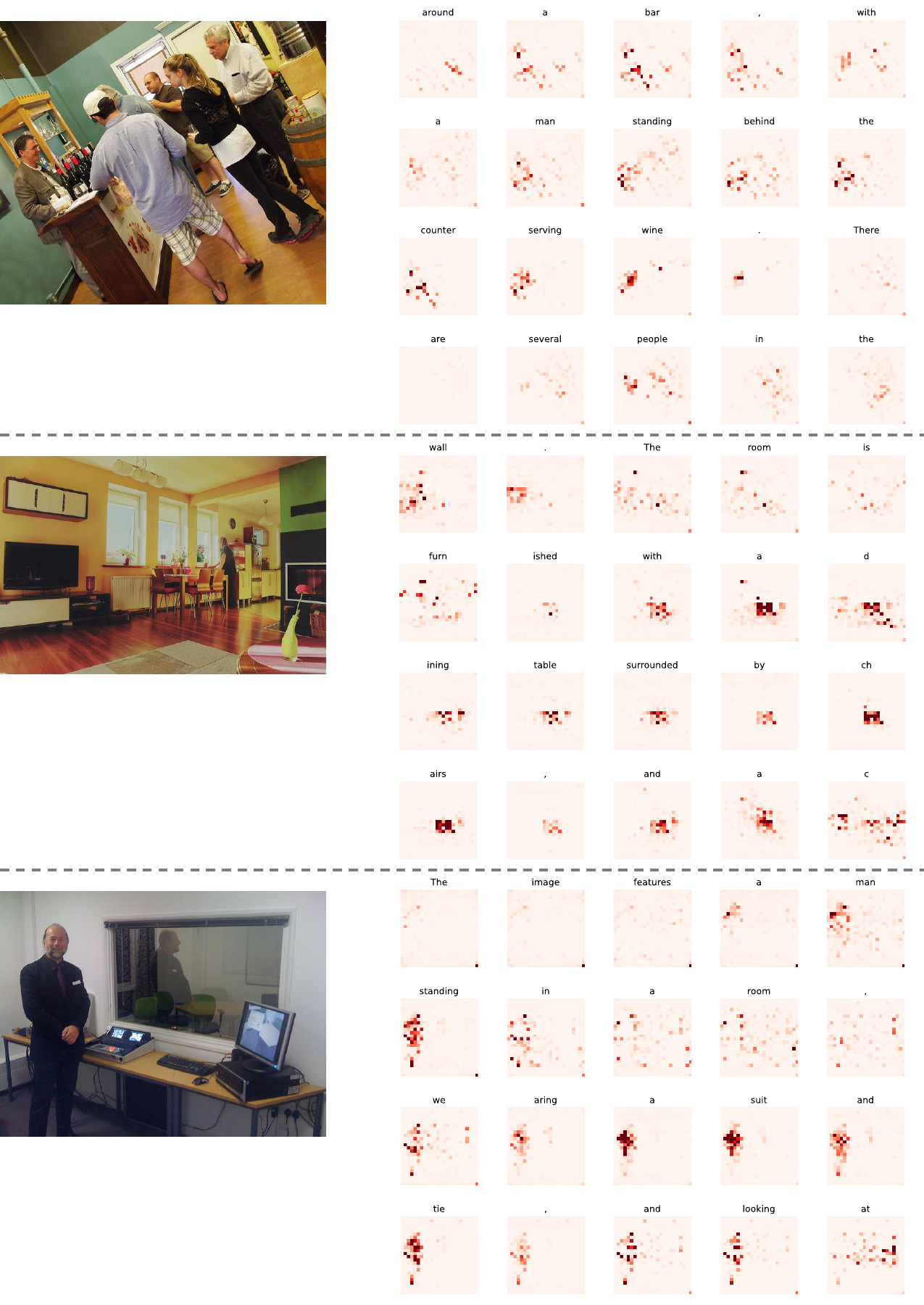}
    \caption{\textbf{Attention maps in complex scenes.} Our approach is able to reason and ground complex phrases (\eg, ``a man standing behind the counter serving wine'') via the attention maps in the LMM. In complex scenes with other similar objects (\eg other men in the first image, another table on the left side of the second image, and the man in the mirror in the third image), our approach can still correctly locate the target object by finding the point with the highest attention value.}
    \label{fig:complex}
\end{figure*}

\begin{figure*}
    \centering
    \includegraphics[width=0.8\textwidth]{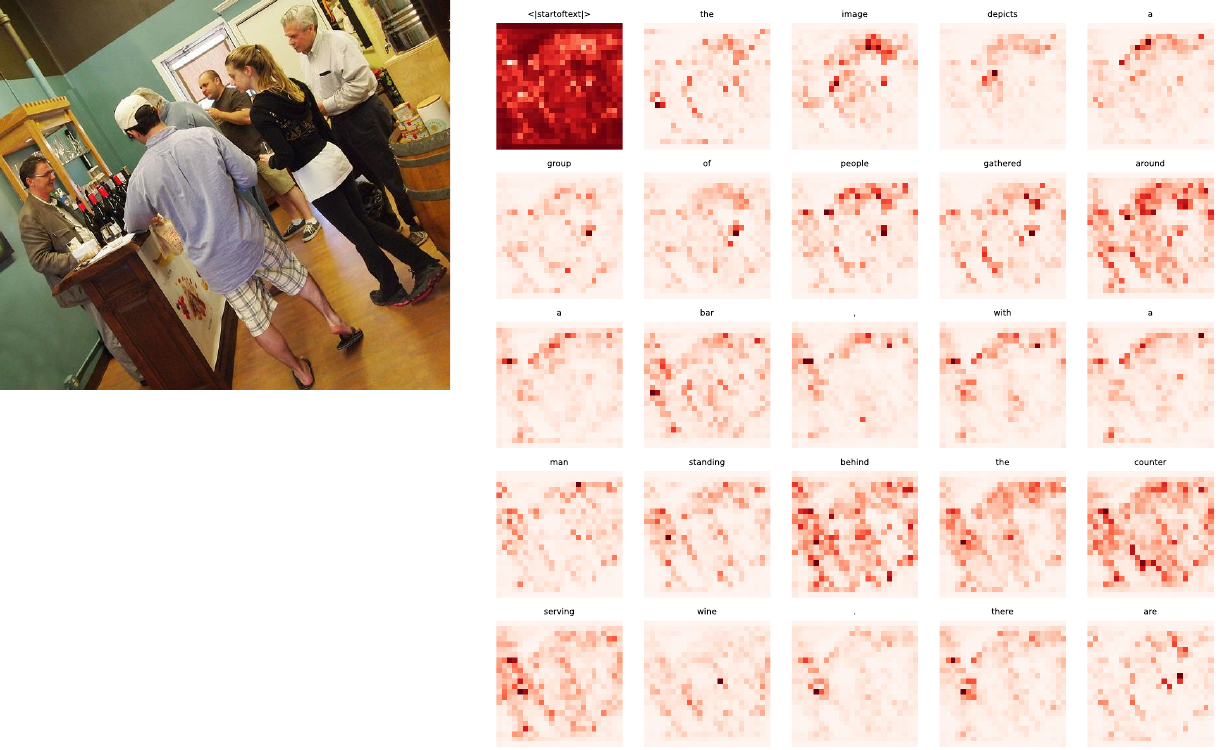}
    \caption{\textbf{Attention maps from the diffusion U-Net.} The U-Net in the diffusion model also computes cross-attention between image patches and a given text condition. However, the attention maps are more noisy compared with the attention maps in the LMM (see Figure~\ref{fig:complex}), and thus are less effective in visual grounding.}
    \label{fig:diff-attn}
\end{figure*}

\noindent\textbf{Visualization of attention maps.} We visualize the attention maps in \ourmodel in complex scenes in Figure~\ref{fig:complex}, to demonstrate the ability to differentiate similar visual entities and understand complex phrases. In Figure~\ref{fig:diff-attn}, we examine the attention maps inside the diffusion U-Net, which uses a CLIP text encoder. Due to the limited capacity of the text encoder, the attention maps produced by the diffusion U-Net are ineffective for visual grounding.

\end{document}